\documentclass[sigconf,nonacm]{acmart}

\setcopyright{none}
\renewcommand\footnotetextcopyrightpermission[1]{}
\usepackage{booktabs} 

\acmConference{}{}{}

\begin{document}
\title{Multiple Classification with Split Learning}

\author{Jongwon Kim}
\affiliation{%
  \institution{Gwangju Institute of Science and Technology}
  \city{Gwangju}
  \country{Korea}
}
\email{jongwonkim@gm.gist.ac.kr}

\author{Sungho Shin}
\affiliation{%
  \institution{Gwangju Institute of Science and Technology}
  \city{Gwangju}
  \country{Korea}
}
\email{hogili89@gm.gist.ac.kr}

\author{Yeonguk Yu}
\affiliation{%
  \institution{Gwangju Institute of Science and Technology}
  \city{Gwangju}
  \country{Korea}
}

\email{yeon_guk@gm.gist.ac.kr}

\author{Junseok Lee}
\affiliation{%
  \institution{Gwangju Institute of Science and Technology}
  \city{Gwangju}
  \country{Korea}
}

\email{junseoklee@gm.gist.ac.kr}

\author{Kyoobin Lee}
\authornote{Correspondence Author}
\affiliation{%
  \institution{Gwangju Institute of Science and Technology}
  \city{Gwangju}
  \country{Korea}}
\email{kyoobinlee@gist.ac.kr}

\renewcommand{\shortauthors}{Jongwon Kim et al.}

\begin{abstract}
Privacy issues were raised in the process of training deep learning in medical, mobility, and other fields. To solve this problem, we present privacy-preserving distributed deep learning method that allow clients to learn a variety of data without direct exposure. We divided a single deep learning architecture into a common extractor, a cloud model and a local classifier for the distributed learning. First, the common extractor, which is used by local clients, extracts secure features from the input data. The secure features also take the role that the cloud model can employ various task and diverse types of data. The feature contain the most important information that helps to proceed various task. Second, the cloud model including most parts of the whole training model gets the embedded features from the massive local clients, and performs most of deep learning operations which takes severe computing cost. 
After the operations in cloud model finished, outputs of the cloud model send back to local clients. 
Finally, the local classifier determined classification results and delivers the results to local clients. 
When clients train models, our model does not directly expose sensitive information to exterior network. 
During the test, the average performance improvement was 2.63\% over the existing local training model. 
However, in a distributed environment, there is a possibility of inversion attack due to exposed features. 
For this reason, we experimented with the common extractor to prevent data restoration. The quality of restoration of the original image was tested by adjusting the depth of the common extractor.
As a result, we found that the deeper the common extractor, the restoration score decreased to 89.74.
\end{abstract}

\newcommand\HH{
  \global\let\savedtextbullet\textbullet
  \gdef\textbullet{%
    \par\noindent\savedtextbullet\global\let\textbullet\savedtextbullet
  }%
}



\maketitle

\section{Introduction}

Deep learning has shown many outstanding performances in many fields such as image analysis \cite{Iizuka2020, Back2020, Moen2019}, signal analysis \cite{Seo2020, Lee2020}, and others. The successes of deep learning  come from the huge amounts of accessible data. Because some data can intrude the privacy of individuals, even some is too sensitive, not all data can be shared for the deep learning. These personal data-related issues cause people to worry about privacy breaches. In order to solve these problems, Shokri et al. present deep learning learning algorithms that communicate between clients and servers without directly exposing data and receive and learn global parameters to ensure privacy of individuals \cite{10.1145/2810103.2813687}. Since the presentation of Federated learning, studies have been conducted on how to incorporate weights learned from clients. Among them, McMahan et al presented the federated averaging algorithm for sharing model weight of each client when conducting federated guidance \cite{pmlr-v54-mcmahan17a}. Also, Zhu and Jin propose modified SET algorithm for weight update of federated learning model \cite{8744465}. To enhance the security of existing federated ordering structures, Bonawitz et al added secure aggregation in the middle to calculate the weights to be updated in other spaces and send them back to the cloud \cite{10.1145/3133956.3133982}. Application studies using Federated learning include the mobile keyboard pre-diction algorithm presented by Hard et al. \cite{hard2018federated} and architecture and applications for the federated learning framework presented by Yang et al \cite{10.1145/3298981}. Federated learning presupposes learning from each client. However, it can be difficult to utilize when clients do not have sufficient computing power. The split learning algorithm, which allows the cloud server to bear most of the computations instead of these clients, was presented by Vepakomma et al \cite{vepakomma2018split}. Further research has presented security issues and solutions arising from Split learning \cite{vepakommareducing}. 

Federated learning and split learning are algorithms that eventually learn the same tasks. On the other hand, Caruana's proposed Multitask learning makes it possible to perform several tasks \cite{caruana1997multitask}. Several studies have been conducted since Multitask learning was presented. Sener et al present multi-task leading algorithm for multi-objective optimization \cite{NIPS2018_7334}. Kendall et al presented an algorithm for learning three different tasks: semantic segmentation, instance segmentation, and depth estimation \cite{Kendall_2018_CVPR}. Liu et al present an end-to-end model for learning segmentation and depth estimation \cite{Liu_2019_CVPR}. Multitask learning has shown some performance, but has the potential for security issues. Multitask leading has shown some performance, but has the potential for security issues. For that reason, the multi-task leading algorithm with federated leading was studied. Smith et al presented MOCHA \cite{NIPS2017_7029}, a multi-task leading algorithm with federated learning, and Corinzia and Buhmann presented a federated multi-task leading algorithm called VIRTURAL \cite{corinzia2019variational}.
 
There are several researches \cite{Fredrikson2015, Yang2019, NIPS2019_9617} to attack the forementioned distributed learning systems. Fredrikson et al. attacked the split learning by reconstruting the input images only using the embedded features, not any other input related information \cite{Fredrikson2015}. Yang et al. and Zhu et al. attacked the federated learning by reconstructing the input images from the gradients, which are flowing through the model during backpropagation. Due to the severe competition between hiding and seeking the private information, fully encrypted deep learning systems are not exist at present. We propose split learning based multiple classification learning, which also can be attacked by inversion attacks. To infer the reconstruction we increased the depth of embedding models, and also increased the proportion of non-linear function in the feature extracting layers.

In this paper, we propose a learning method in which personal privacy is guaranteed and information can be shared with other clients while also utilizing the resources of the cloud. Our model is largely divided into three parts. Two parts are computed on a local computer, and only the other part is computed on a cloud server. 
On the local computer, common extractor and local classifier are trained. 
Unlike the existing distributed learning method, our model receives input data with different types and forms for various task. 
So, our model makes standardized features through common extractor. 
In contrast, cloud models are trained in cloud servers. The cloud model receives the standardized feature from the common extractor, than the cloud model outputs processed feature. 
Finally, The processed feature sent to local classifier.
In this distributed environment which split local part and cloud part, only forward feature and backward gradient values are transmitted and received, which keep our system secured from malicious attackers. 
We assumed 3 clients and used Cifar10 to learn the model. Each client will perform model learning through cloud computing without exposing the original data through the built model. For this reason, each client's original data can be studied without being exposed directly.

\begin{figure*}
\includegraphics[scale=0.48]{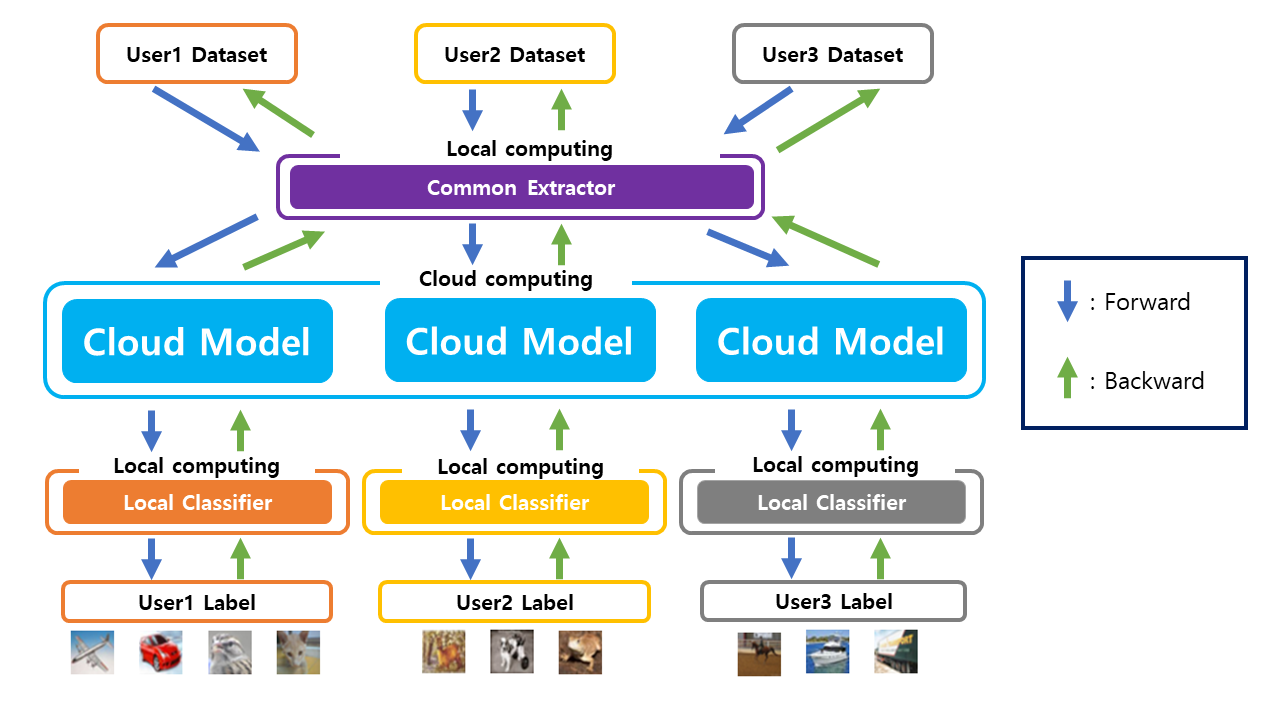}
\caption{Distributed deep learning model structure. This model is divided into the local client compute part and the cloud server compute part. Therefore, the original training data is not exposed to the outside world, and only forward features and backward gradient are sent and received.}
\label{fig:inversion}
\end{figure*}

\section{Related work}

\subsection{Distributed deep learning}
Federated learning has been proposed to address privacy issues arising from data learning \cite{10.1145/2810103.2813687}. Federated learning is a learning method in which personal data with privacy problems are learned in each individual's local computer, collected learning results from cloud servers, and then re-distributed them into a single model. At this time, how to combine the models of the local client has a significant impact on performance. For this reason, studies have been conducted on whether to increase the performance of a single combined model by integrating the model weight or gradient values for each local \cite{pmlr-v54-mcmahan17a, Zhu2020, Wang2020, Duan2019, Mohri2019}.

Federated learning is a great advantage that it can contribute to model learning without exposing sensitive privacy data of individuals. However, the problem exists that local clients should use only the performance of their personal computers entirely. Split learning was proposed to reduce the burden of computing on these individuals \cite{vepakomma2018split}. Split learning is a method of performing part of the model's training operations on cloud servers. Therefore, split learning reduces the burden of computing on local clients and allows them to train heavy models efficiently. However, because part of the training is performed on cloud servers, split learning has the potential to expose relevant information during training. Therefore, an algorithm was presented to prevent the restoration of the information using the gradient value or feature information transmitted \cite{vepakommareducing}.

\subsection{Multi-task learning}
Most models are trained to perform only one task. Multitask leading is presented to overcome these limitations \cite{caruana1997multitask}. Multitask learning is mainly used in two ways: hard parameter sharing and soft parameter sharing \cite{Ruder2017}. The hard parameter sharing method learns about different tasks, but shares the same weight in the beginning and later has different weights. Conversely, the soft parameter sharing method only shares training information with different weights. In addition, Deep Relationship Networks \cite{Long2017}, Full-Adaptive Feature Sharing \cite{Lu2017}, Cross-stitch Networks \cite{Misra2016}, Low Supervision \cite{Sagaard2016}, A Point Many-Task Model \cite{Hashimoto2017}, Weighting loss with certification \cite{Kendall_2018_CVPR}, Tensor factory for MTL \cite{Yang2017}, and Fluice Networks \cite{Ruder2019} training methods exist.

\subsection{Inversion attack}
Numerous methods for privacy preserving deep learning (PPDL) have been proposed \cite{10.1145/2810103.2813687, pmlr-v54-mcmahan17a, vepakomma2018split}. Most widely used methods such as Federated learning \cite{10.1145/2810103.2813687, pmlr-v54-mcmahan17a} and split learning \cite{vepakomma2018split} don't share raw data for the privacy, though training deep learning model requires input data. Instead, embedded features or gradients, which were believed as safe, are shared to train deep learning models. However, Fredrikson et al. developed an inversion attack which reconstruct an input image from the embedded features \cite{Fredrikson2015}. Simply placing decoder models, which have inverted structures of trained model, to embedded features reconstruct input images by optimizing decoder model's parameters can successfully inverting the training model's parameters. Furthermore, Yang et al. reconstruct input images only using softmax results. These results indicate sharing embedded features or softmax results is still exposed to inversion attacks \cite{Yang2019}. Because split learning \cite{vepakomma2018split} shared embedded features to the cloud server, the split learning is still exposed to reconstruction attacks. Zhu et al. attacks federated learning by reconstructing the input images only from leaked gradients during back-propagation process \cite{NIPS2019_9617}. To reconstruct input image, attacker place gaussian noise as parameters. The attackers optimized the gaussian noises to become similar with gradients leaked from the original training processes. After the optimization, the optimized parameters show extremely similar figures with the original image, which poses severe danger in federated learning. 
 Due to the inversion attacks, distributed deep learning systems needs to pay attention to developing reliable process for training and validating deep learning models. Our research relieve the reconstruction danger by making feature extracting layers deeper by adding non-linear functions such as convolution, max pooling and rectified linear unit (ReLU).

\section{Method}
\subsection{Distributed deep learning}

The distributed deep learning system was built with one cloud server, more than two local clients. The cloud server will be the largest part of the model training. And the local trains the rest of the model that except the cloud server's part. so the computational burden is relatively low. The distributed deep learning system is largely divided into Common Extractor, Cloud Model, and Local Classifier. 
First, the entire local client will share a neural network called common extractor. Therefore, local clients share the weights of some of the training models. 
This shared neural network is learned knowledge about various feature. For this reason, the common extractor analyze important information more than non-shared neural network.
Second, The cloud server is responsible for most model training except common extractor and classifier. In other words, the cloud server will cover a large operation in overall model learning, and the client will have a relatively small operation. If clients want to pay attention to privacy, they can increase the extent to which the common extractor pays for the entire model. However, this entails an increase in the computational burden that must be borne locally. Finally, the local classifier includes the model's last fully connected layer.

Local clients who want to learn through cloud servers send training requests to the cloud. After local clients request training to the server, each client inputs their own training data into the same initialized common extractor. 
The common extractor outputs features of the input data and then sends them to the cloud model. 
Since clients use the common extractor, the cloud model receives feature with standardized shape from the common extractor and send a cloud model feature to local classifier as output.
Finally, the result from classifier is compared with ground true to obtain the loss value. 
After this forward propagation, backpropagation is performed through the loss value. 
The backpropagation direction is reversed by forward propagation and transmits the gradient value.

When the entire local client has completed one epoch training, obtain the average weight for the common extractor of the local clients which have learned with different data. Then update all common extractors to the average weight. After the weight of the common extractor is updated, the models of the local client and cloud server continue the training again.

\subsection{Inversion attack using embedded features}

\begin{figure*}
\includegraphics[scale=0.65]{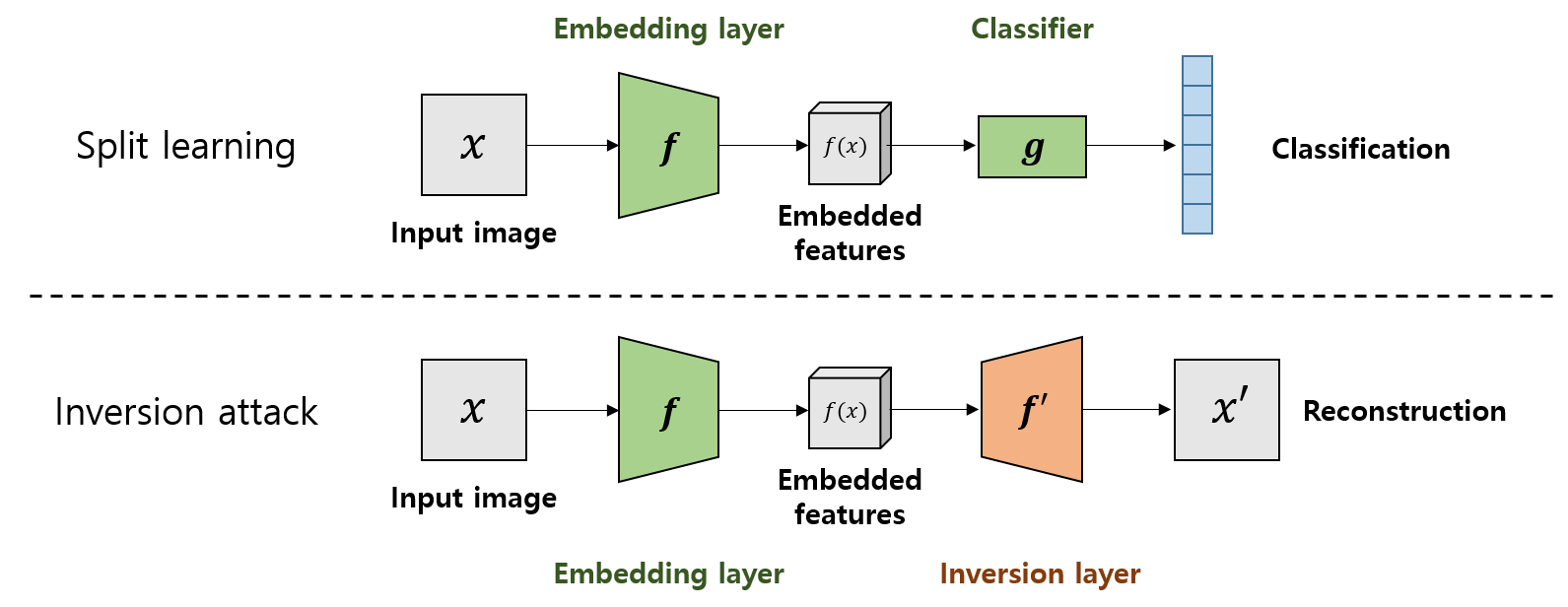}
\caption{Outline of the inversion attacks which applied to the split learning.}
\label{fig:inversion}
\end{figure*}

Inversion attacks reconstruct input image only using the partial information such as embedded features or gradient flows. Because our distributed deep learning system follow the split learning structures \cite{vepakomma2018split}, reconstructing input images from the embedded features can be problematic. Inversion attack proposed by Yang et al. reconstructed input images from the embedded features successfully by simply adding the decoder model, which has inverted structures of original embedding layers (Fig. \ref{fig:inversion}) \cite{Yang2019}.

\begin{equation}
  y = g(f(x)), \quad x\in R^3
\label{eq:classification}
\end{equation}
\begin{equation}
  x' = f'(f(x)), \quad x\in R^3
\label{eq:inversion}
\end{equation}

 In Fig. \ref{eq:classification} show split learning process for training classification model. Input images pass through the embedding layers $f(\cdot)$, and the embedded features $f(x)$ are shared with the centralized server for training classifier $g(\cdot)$. However, when the attackers attach decoder $f'(\cdot)$ to the embedded features, input image can be reconstructed. The decoder has inverted structures of embedding layers, and through the repetitive training of decoder embedded features can be inverted to input images successfully. For example, convolution layer in embedding layers can be inverted with the transpose convolution layer, and ReLU activation function can be inverted with tanh function.

\section{Experiment}
\subsection{Distributed deep learning}
\subsubsection{Experimental design}\hspace{\fill}\\
\noindent
To evaluate the performance of the distributed learning model, We used the mnist, cifar10 and imagenet dataset as training data for distributed environment models. Since we assumed three local clients, we proceeded by distributing the data to each local client. In this experiment, we distributed class data from 0 to 3 to local client1, class data from 4 to 6 to local client2, and class data from 7 to 9 to local client3.

And for the training model, the distributed learning model was built by partitioning the ResNet50. To check the impact on the depth of the common extractor, we proceeded with the training by changing the depth of the common extractor. In this experiment, we set common extractor from input to layer1 in experiment1, from input to layer2 in experiment 2, and from input to layer3 in experiment 3.

We wrote the python code for the experiment using pytorch. Different training parameters were used for each data training. First of all, when using mnist as training data, the epoch size is 20, batch number is 200, and the leading rate is 0.001. Secondly, when cifar10 is used as training data, the epoch size is 200 and the batch number is 200 is 0.1. Thirdly, when the imagenet is used as training data, the epoch size is 200, batch number is 64, and the leading rate is 0.01. On the other hand, optimizer and loss function were used the same as SGD optimizer and cross entropy.

 \begin{figure*}
\includegraphics[scale=0.65]{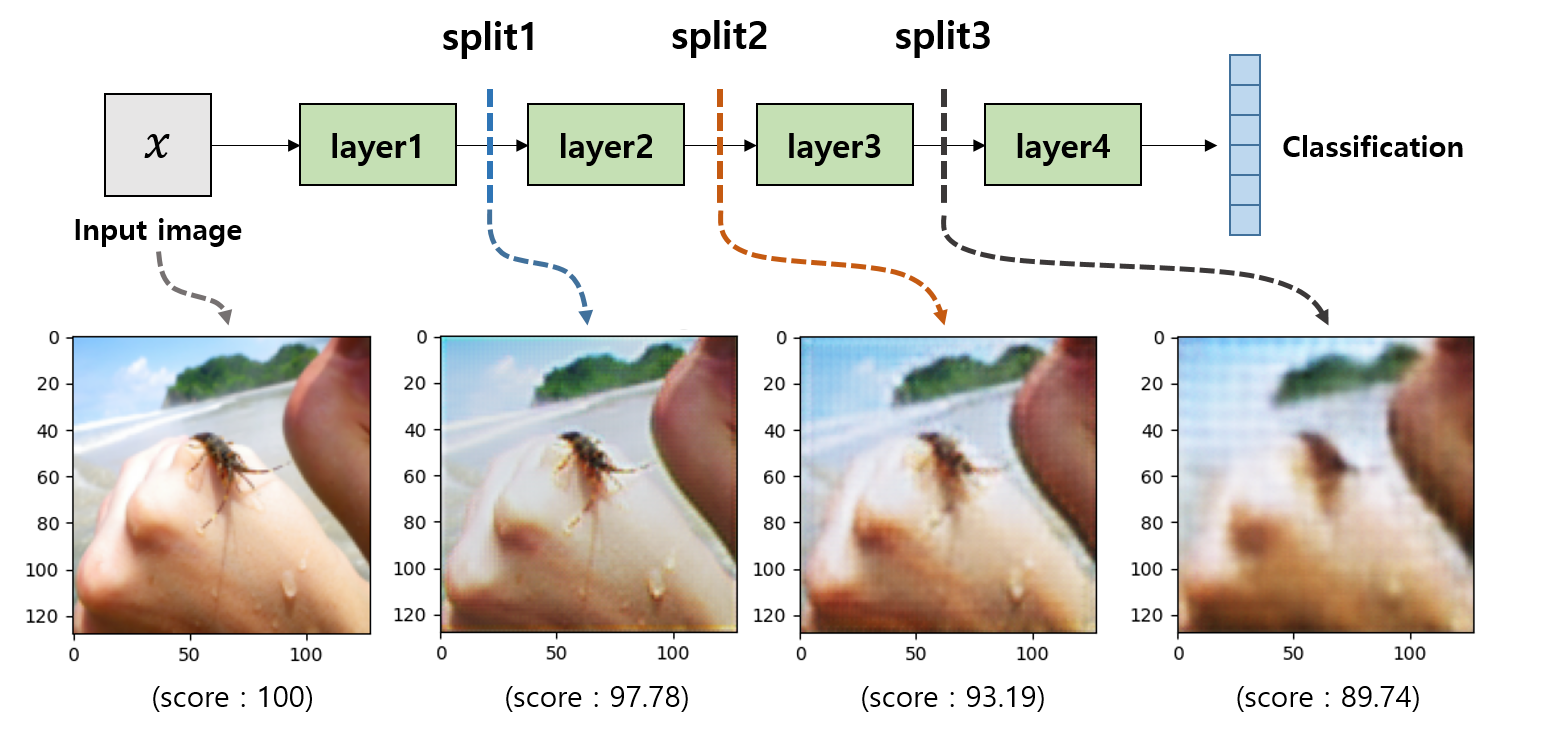}
\caption{The reconstructed results of inversion attack depending on depth in split learning.}
\label{fig:inversion_result}
\end{figure*}

 \begin{table}[]
     \centering
     \caption{Distributed Learning and Non-Distributed learning test results. Sharing means sharing part of the model as a common extractor. On the contrary, non-sharing means not sharing part of the model.}
     \begin{tabular}{c|c|c|c}
          \multicolumn{1}{c|}{Dataset}   & \multicolumn{1}{c|}{Client} & \multicolumn{1}{c|}{\begin{tabular}[c]{@{}c@{}}Accuracy(\%)\\ (sharing)\end{tabular}} & \multicolumn{1}{c}{\begin{tabular}[c]{@{}c@{}}Accuracy(\%)\\ (non-sharing)\end{tabular}}\\\hline
                     & Client1 & 99.18 & 98.84\\
            Mnist    & Client2 & 99.22 & 98.62\\
                     & Client3 & 98.57 & 98.60\\\hline
                     & Client1 & 92.97 & 92.02\\
            Cifar10  & Client2 & 94.50 & 92.20\\
                     & Client3 & 97.83 & 95.40\\\hline
                     & Client1 & 71.50 & 70.50\\
            Imagenet & Client2 & 81.33 & 76.00\\
                     & Client3 & 87.33 & 80.66\\\hline
     \end{tabular}
     \label{tab:my_label}
 \end{table}

\subsubsection{Distributed model performance result}\hspace{\fill}\\
Table 1 shows the experimental results. The results in the table compared the distributed ordering model that shared weight with the model that did not share weight. For the mnist dataset, the overall client-specific performance showed slightly higher performance, although not large when the weight was shared. In cifiar10 dataset, the accuracy of models that shared Weight and those that did not share it was not significantly different. Similarly two datasets, the sharing model had higher performance for all clients in the case of the Imagenet dataset.

Overall, there was no significant difference in accuracy depending on whether weight was shared or not. Based on this, it was assumed that simply averaging the weights was not effective. In other words, the performance of the distributed learning model will depend on how the weight of the common extractor is shared.

\subsection{Inversion attack to common extractor}
\subsubsection{Experimental design}\hspace{\fill}\\
We examine the effects of common extractor's depth on reconstruction success. The measure of success is defined as the differences of pixels between input and reconstructed images, in Eq. \ref{eq:reconstruction}. When the differences between input and reconstructed images are small, then the reconstruction score indicate relatively high value.

\begin{equation}
  Reconstruction Score = 1 - 1000 * (\sum_{i=1}^{N}(x_i - x_i')^2)/N, \quad x_i\in R^3
\label{eq:reconstruction}
 \end{equation}

 $x$ indicates original image, $x'$ indicates reconstructed image, and $N$ means the number of train samples. Using the reconstruction score, we compared the reconstruction score when the depth of common extractors are differ. ResNet50 consists of four main layers and one classifier. When splitting the resnet50 into two parts, common extractor and others, we controlled the divided place, which indicates depth of common extractor. For three common extractors, which have different depth (layer1, layer2, and layer3 in Fig. \ref{fig:inversion_result}), reconstruction scores are compared.

\subsubsection{Reconstruction results depending on layers' depth}\hspace{\fill}\\
When the depth of common extractor increased, the reconstruction score decreased in Fig. \ref{fig:inversion_result}. Non-linear function such as ReLU, convolution, and max pooling compressed the image information by removing less informative features. Increased depth means the increased number of non-linear function, which is irreversible to removing huge amounts of information. In Fig. \ref{fig:inversion_result}, the reconstruction score decreased almost linearly when the depth of common extractor increased. Though the reconstruction score decreased, the reconstructed result in split3 can be recognized as hands, insects, mountain, and sea in background. Therefore, additional privacy-preserving method like differential privacy is required to be merged with our systems, which are our future plans.

\section{Conclusion}
In this experiment, we proposed a distribution deep learning model to solve the privacy problem and the resource problem of local clients. The model divides the training operation to be in charge of the local client and the cloud server, and limits the transmitted and received data to the forward feature and backward gradient. Therefore, it is possible to utilize the resources of the cloud without directly exposing the original data. To verify the model's performance, we conducted a test using mnist, cifar10, and imagenet data sets. As a result, similar accuracy was obtained without significant gap in performance. Also, we experimented with how the layer depth of the common extractor affects information security. As a result, it was confirmed that it became more difficult to recover information as the layer depth became deeper. For the more reliable privacy preserving system, we further research is to apply differential privacy in our system.

\begin{acks}
 This work was supported by Institute of Information \& communications Technology Planning \& Evaluation (IITP) grant funded by the Korea government(MSIT) (No. 2020-0-00857, Development of cloud robot intelligence augmentation, sharing and framework technology to integrate and enhance the intelligence of multiple robots).

\end{acks}

\bibliographystyle{ACM_Reference_Format}
\bibliography{reference}


\begin{thebibliography}{34}


\ifx \showCODEN    \undefined \def \showCODEN     #1{\unskip}     \fi
\ifx \showDOI      \undefined \def \showDOI       #1{#1}\fi
\ifx \showISBNx    \undefined \def \showISBNx     #1{\unskip}     \fi
\ifx \showISBNxiii \undefined \def \showISBNxiii  #1{\unskip}     \fi
\ifx \showISSN     \undefined \def \showISSN      #1{\unskip}     \fi
\ifx \showLCCN     \undefined \def \showLCCN      #1{\unskip}     \fi
\ifx \shownote     \undefined \def \shownote      #1{#1}          \fi
\ifx \showarticletitle \undefined \def \showarticletitle #1{#1}   \fi
\ifx \showURL      \undefined \def \showURL       {\relax}        \fi
\providecommand\bibfield[2]{#2}
\providecommand\bibinfo[2]{#2}
\providecommand\natexlab[1]{#1}
\providecommand\showeprint[2][]{arXiv:#2}

\bibitem[\protect\citeauthoryear{Back, Kim, Kang, Choi, and Lee}{Back
  et~al\mbox{.}}{2020}]%
        {Back2020}
\bibfield{author}{\bibinfo{person}{Seunghyeok Back}, \bibinfo{person}{Jongwon
  Kim}, \bibinfo{person}{Raeyong Kang}, \bibinfo{person}{Seungjun Choi}, {and}
  \bibinfo{person}{Kyoobin Lee}.} \bibinfo{year}{2020}\natexlab{}.
\newblock \showarticletitle{{Segmenting Unseen Industrial Components in a Heavy
  Clutter Using RGB-D Fusion and Synthetic Data}}.
\newblock  (\bibinfo{date}{feb} \bibinfo{year}{2020}).
\newblock
\showeprint[arxiv]{2002.03501}
\showURL{%
\url{http://arxiv.org/abs/2002.03501}}


\bibitem[\protect\citeauthoryear{Bonawitz, Ivanov, Kreuter, Marcedone, McMahan,
  Patel, Ramage, Segal, and Seth}{Bonawitz et~al\mbox{.}}{2017}]%
        {10.1145/3133956.3133982}
\bibfield{author}{\bibinfo{person}{Keith Bonawitz}, \bibinfo{person}{Vladimir
  Ivanov}, \bibinfo{person}{Ben Kreuter}, \bibinfo{person}{Antonio Marcedone},
  \bibinfo{person}{H.~Brendan McMahan}, \bibinfo{person}{Sarvar Patel},
  \bibinfo{person}{Daniel Ramage}, \bibinfo{person}{Aaron Segal}, {and}
  \bibinfo{person}{Karn Seth}.} \bibinfo{year}{2017}\natexlab{}.
\newblock \showarticletitle{{Practical secure aggregation for
  privacy-preserving machine learning}}. In \bibinfo{booktitle}{{\em
  Proceedings of the ACM Conference on Computer and Communications Security}}.
  \bibinfo{publisher}{Association for Computing Machinery},
  \bibinfo{address}{New York, NY, USA}, \bibinfo{pages}{1175--1191}.
\newblock
\showISBNx{9781450349468}
\showISSN{15437221}
\showDOI{%
\url{https://doi.org/10.1145/3133956.3133982}}


\bibitem[\protect\citeauthoryear{{Brendan McMahan}, Moore, Ramage, Hampson, and
  {Ag{\"{u}}era y Arcas}}{{Brendan McMahan} et~al\mbox{.}}{2017}]%
        {pmlr-v54-mcmahan17a}
\bibfield{author}{\bibinfo{person}{H. {Brendan McMahan}},
  \bibinfo{person}{Eider Moore}, \bibinfo{person}{Daniel Ramage},
  \bibinfo{person}{Seth Hampson}, {and} \bibinfo{person}{Blaise {Ag{\"{u}}era y
  Arcas}}.} \bibinfo{year}{2017}\natexlab{}.
\newblock \showarticletitle{{Communication-efficient learning of deep networks
  from decentralized data}}. In \bibinfo{booktitle}{{\em Proceedings of the
  20th International Conference on Artificial Intelligence and Statistics,
  AISTATS 2017}}. \bibinfo{publisher}{PMLR}, \bibinfo{address}{Fort Lauderdale,
  FL, USA}, \bibinfo{pages}{1273--1282}.
\newblock
\showISSN{2640-3498}
\showeprint{1602.05629}
\showURL{%
\url{http://proceedings.mlr.press/v54/mcmahan17a.html}}


\bibitem[\protect\citeauthoryear{Caruana}{Caruana}{1997}]%
        {caruana1997multitask}
\bibfield{author}{\bibinfo{person}{Rich Caruana}.}
  \bibinfo{year}{1997}\natexlab{}.
\newblock \showarticletitle{{Multitask Learning}}.
\newblock \bibinfo{journal}{{\em Machine Learning\/}} \bibinfo{volume}{28},
  \bibinfo{number}{1} (\bibinfo{year}{1997}), \bibinfo{pages}{41--75}.
\newblock
\showISSN{08856125}
\showDOI{%
\url{https://doi.org/10.1023/A:1007379606734}}


\bibitem[\protect\citeauthoryear{Corinzia and Buhmann}{Corinzia and
  Buhmann}{2019}]%
        {corinzia2019variational}
\bibfield{author}{\bibinfo{person}{Luca Corinzia} {and}
  \bibinfo{person}{Joachim~M. Buhmann}.} \bibinfo{year}{2019}\natexlab{}.
\newblock \showarticletitle{{Variational Federated Multi-Task Learning}}.
\newblock  (\bibinfo{date}{jun} \bibinfo{year}{2019}).
\newblock
\showeprint[arxiv]{1906.06268}
\showURL{%
\url{http://arxiv.org/abs/1906.06268}}


\bibitem[\protect\citeauthoryear{Duan, Liu, Chen, Tan, Ren, Qiao, and
  Liang}{Duan et~al\mbox{.}}{2019}]%
        {Duan2019}
\bibfield{author}{\bibinfo{person}{Moming Duan}, \bibinfo{person}{Duo Liu},
  \bibinfo{person}{Xianzhang Chen}, \bibinfo{person}{Yujuan Tan},
  \bibinfo{person}{Jinting Ren}, \bibinfo{person}{Lei Qiao}, {and}
  \bibinfo{person}{Liang Liang}.} \bibinfo{year}{2019}\natexlab{}.
\newblock \showarticletitle{{Astraea: Self-balancing federated learning for
  improving classification accuracy of mobile deep learning applications}}. In
  \bibinfo{booktitle}{{\em Proceedings - 2019 IEEE International Conference on
  Computer Design, ICCD 2019}}. \bibinfo{publisher}{Institute of Electrical and
  Electronics Engineers Inc.}, \bibinfo{pages}{246--254}.
\newblock
\showISBNx{9781538666487}
\showDOI{%
\url{https://doi.org/10.1109/ICCD46524.2019.00038}}
\showeprint[arxiv]{1907.01132}


\bibitem[\protect\citeauthoryear{Fredrikson, Jha, and Ristenpart}{Fredrikson
  et~al\mbox{.}}{2015}]%
        {Fredrikson2015}
\bibfield{author}{\bibinfo{person}{Matt Fredrikson}, \bibinfo{person}{Somesh
  Jha}, {and} \bibinfo{person}{Thomas Ristenpart}.}
  \bibinfo{year}{2015}\natexlab{}.
\newblock \showarticletitle{{Model inversion attacks that exploit confidence
  information and basic countermeasures}}. In \bibinfo{booktitle}{{\em
  Proceedings of the ACM Conference on Computer and Communications Security}},
  Vol.~\bibinfo{volume}{2015-Octob}. \bibinfo{publisher}{Association for
  Computing Machinery}, \bibinfo{address}{New York, New York, USA},
  \bibinfo{pages}{1322--1333}.
\newblock
\showISBNx{9781450338325}
\showISSN{15437221}
\showDOI{%
\url{https://doi.org/10.1145/2810103.2813677}}


\bibitem[\protect\citeauthoryear{Hard, Rao, Mathews, Ramaswamy, Beaufays,
  Augenstein, Eichner, Kiddon, and Ramage}{Hard et~al\mbox{.}}{2018}]%
        {hard2018federated}
\bibfield{author}{\bibinfo{person}{Andrew Hard}, \bibinfo{person}{Kanishka
  Rao}, \bibinfo{person}{Rajiv Mathews}, \bibinfo{person}{Swaroop Ramaswamy},
  \bibinfo{person}{Fran{\c{c}}oise Beaufays}, \bibinfo{person}{Sean
  Augenstein}, \bibinfo{person}{Hubert Eichner}, \bibinfo{person}{Chlo{\'{e}}
  Kiddon}, {and} \bibinfo{person}{Daniel Ramage}.}
  \bibinfo{year}{2018}\natexlab{}.
\newblock \showarticletitle{{Federated Learning for Mobile Keyboard
  Prediction}}.
\newblock  (\bibinfo{date}{nov} \bibinfo{year}{2018}).
\newblock
\showeprint[arxiv]{1811.03604}
\showURL{%
\url{http://arxiv.org/abs/1811.03604}}


\bibitem[\protect\citeauthoryear{Hashimoto, Xiong, Tsuruoka, and
  Socher}{Hashimoto et~al\mbox{.}}{2017}]%
        {Hashimoto2017}
\bibfield{author}{\bibinfo{person}{Kazuma Hashimoto}, \bibinfo{person}{Caiming
  Xiong}, \bibinfo{person}{Yoshimasa Tsuruoka}, {and} \bibinfo{person}{Richard
  Socher}.} \bibinfo{year}{2017}\natexlab{}.
\newblock \showarticletitle{{A joint many-task model: Growing a neural network
  for multiple NLP tasks}}. In \bibinfo{booktitle}{{\em EMNLP 2017 - Conference
  on Empirical Methods in Natural Language Processing, Proceedings}}.
  \bibinfo{publisher}{Association for Computational Linguistics},
  \bibinfo{pages}{1923--1933}.
\newblock
\showISBNx{9781945626838}
\showeprint[arxiv]{1611.01587}
\showURL{%
\url{http://arxiv.org/abs/1611.01587}}


\bibitem[\protect\citeauthoryear{Iizuka, Kanavati, Kato, Rambeau, Arihiro, and
  Tsuneki}{Iizuka et~al\mbox{.}}{2020}]%
        {Iizuka2020}
\bibfield{author}{\bibinfo{person}{Osamu Iizuka}, \bibinfo{person}{Fahdi
  Kanavati}, \bibinfo{person}{Kei Kato}, \bibinfo{person}{Michael Rambeau},
  \bibinfo{person}{Koji Arihiro}, {and} \bibinfo{person}{Masayuki Tsuneki}.}
  \bibinfo{year}{2020}\natexlab{}.
\newblock \showarticletitle{{Deep Learning Models for Histopathological
  Classification of Gastric and Colonic Epithelial Tumours}}.
\newblock \bibinfo{journal}{{\em Scientific Reports\/}} \bibinfo{volume}{10},
  \bibinfo{number}{1} (\bibinfo{date}{dec} \bibinfo{year}{2020}),
  \bibinfo{pages}{1--11}.
\newblock
\showISSN{20452322}
\showDOI{%
\url{https://doi.org/10.1038/s41598-020-58467-9}}


\bibitem[\protect\citeauthoryear{Kendall, Gal, and Cipolla}{Kendall
  et~al\mbox{.}}{2018}]%
        {Kendall_2018_CVPR}
\bibfield{author}{\bibinfo{person}{Alex Kendall}, \bibinfo{person}{Yarin Gal},
  {and} \bibinfo{person}{Roberto Cipolla}.} \bibinfo{year}{2018}\natexlab{}.
\newblock \showarticletitle{{Multi-task Learning Using Uncertainty to Weigh
  Losses for Scene Geometry and Semantics}}. In \bibinfo{booktitle}{{\em
  Proceedings of the IEEE Computer Society Conference on Computer Vision and
  Pattern Recognition}}. \bibinfo{publisher}{IEEE},
  \bibinfo{pages}{7482--7491}.
\newblock
\showISBNx{9781538664209}
\showISSN{10636919}
\showDOI{%
\url{https://doi.org/10.1109/CVPR.2018.00781}}
\showeprint[arxiv]{1705.07115}


\bibitem[\protect\citeauthoryear{Lee, Shin, Kim, Thien, Lee, and Kim}{Lee
  et~al\mbox{.}}{2020}]%
        {Lee2020}
\bibfield{author}{\bibinfo{person}{Youngjoo Lee}, \bibinfo{person}{Sungho
  Shin}, \bibinfo{person}{Sungchul Kim}, \bibinfo{person}{Nguyen Thien},
  \bibinfo{person}{Kyoobin Lee}, {and} \bibinfo{person}{Jae~Gwan Kim}.}
  \bibinfo{year}{2020}\natexlab{}.
\newblock \showarticletitle{{Classification of meat freshness based on deep
  learning using data from diffuse reflectance spectroscopy (Conference
  Presentation)}}. In \bibinfo{booktitle}{{\em Imaging, Manipulation, and
  Analysis of Biomolecules, Cells, and Tissues XVIII}},
  \bibfield{editor}{\bibinfo{person}{Daniel~L. Farkas},
  \bibinfo{person}{James~F. Leary}, {and} \bibinfo{person}{Attila Tarnok}}
  (Eds.), Vol.~\bibinfo{volume}{11243}. \bibinfo{publisher}{SPIE},
  \bibinfo{pages}{50}.
\newblock
\showISBNx{9781510632493}
\showDOI{%
\url{https://doi.org/10.1117/12.2545967}}


\bibitem[\protect\citeauthoryear{Liu, Johns, and Davison}{Liu
  et~al\mbox{.}}{2019}]%
        {Liu_2019_CVPR}
\bibfield{author}{\bibinfo{person}{Shikun Liu}, \bibinfo{person}{Edward Johns},
  {and} \bibinfo{person}{Andrew~J Davison}.} \bibinfo{year}{2019}\natexlab{}.
\newblock \showarticletitle{{End-to-end multi-task learning with attention}}.
  In \bibinfo{booktitle}{{\em Proceedings of the IEEE Computer Society
  Conference on Computer Vision and Pattern Recognition}},
  Vol.~\bibinfo{volume}{2019-June}. \bibinfo{pages}{1871--1880}.
\newblock
\showISBNx{9781728132938}
\showISSN{10636919}
\showDOI{%
\url{https://doi.org/10.1109/CVPR.2019.00197}}


\bibitem[\protect\citeauthoryear{Long, Cao, Wang, and Yu}{Long
  et~al\mbox{.}}{2017}]%
        {Long2017}
\bibfield{author}{\bibinfo{person}{Mingsheng Long}, \bibinfo{person}{Zhangjie
  Cao}, \bibinfo{person}{Jianmin Wang}, {and} \bibinfo{person}{Philip~S Yu}.}
  \bibinfo{year}{2017}\natexlab{}.
\newblock \showarticletitle{{Learning multiple tasks with multilinear
  relationship networks}}. In \bibinfo{booktitle}{{\em Advances in Neural
  Information Processing Systems}}, Vol.~\bibinfo{volume}{2017-Decem}.
  \bibinfo{pages}{1595--1604}.
\newblock
\showISSN{10495258}
\showeprint[arxiv]{1506.02117}


\bibitem[\protect\citeauthoryear{Lu, Kumar, Zhai, Cheng, Javidi, and Feris}{Lu
  et~al\mbox{.}}{2017}]%
        {Lu2017}
\bibfield{author}{\bibinfo{person}{Yongxi Lu}, \bibinfo{person}{Abhishek
  Kumar}, \bibinfo{person}{Shuangfei Zhai}, \bibinfo{person}{Yu Cheng},
  \bibinfo{person}{Tara Javidi}, {and} \bibinfo{person}{Rogerio Feris}.}
  \bibinfo{year}{2017}\natexlab{}.
\newblock \showarticletitle{{Fully-adaptive feature sharing in multi-task
  networks with applications in person attribute classification}}. In
  \bibinfo{booktitle}{{\em Proceedings - 30th IEEE Conference on Computer
  Vision and Pattern Recognition, CVPR 2017}},
  Vol.~\bibinfo{volume}{2017-Janua}. \bibinfo{pages}{1131--1140}.
\newblock
\showISBNx{9781538604571}
\showDOI{%
\url{https://doi.org/10.1109/CVPR.2017.126}}
\showeprint[arxiv]{1611.05377}


\bibitem[\protect\citeauthoryear{Misra, Shrivastava, Gupta, and Hebert}{Misra
  et~al\mbox{.}}{2016}]%
        {Misra2016}
\bibfield{author}{\bibinfo{person}{Ishan Misra}, \bibinfo{person}{Abhinav
  Shrivastava}, \bibinfo{person}{Abhinav Gupta}, {and} \bibinfo{person}{Martial
  Hebert}.} \bibinfo{year}{2016}\natexlab{}.
\newblock \showarticletitle{{Cross-Stitch Networks for Multi-task Learning}}.
  In \bibinfo{booktitle}{{\em Proceedings of the IEEE Computer Society
  Conference on Computer Vision and Pattern Recognition}},
  Vol.~\bibinfo{volume}{2016-Decem}. \bibinfo{publisher}{IEEE Computer
  Society}, \bibinfo{pages}{3994--4003}.
\newblock
\showISBNx{9781467388504}
\showISSN{10636919}
\showDOI{%
\url{https://doi.org/10.1109/CVPR.2016.433}}
\showeprint[arxiv]{1604.03539}


\bibitem[\protect\citeauthoryear{Moen, Bannon, Kudo, Graf, Covert, and {Van
  Valen}}{Moen et~al\mbox{.}}{2019}]%
        {Moen2019}
\bibfield{author}{\bibinfo{person}{Erick Moen}, \bibinfo{person}{Dylan Bannon},
  \bibinfo{person}{Takamasa Kudo}, \bibinfo{person}{William Graf},
  \bibinfo{person}{Markus Covert}, {and} \bibinfo{person}{David {Van Valen}}.}
  \bibinfo{year}{2019}\natexlab{}.
\newblock \bibinfo{title}{{Deep learning for cellular image analysis}}.
\newblock   (\bibinfo{date}{dec} \bibinfo{year}{2019}),
  \bibinfo{numpages}{1233--1246}~pages.
\newblock
\showISSN{15487105}
\showDOI{%
\url{https://doi.org/10.1038/s41592-019-0403-1}}


\bibitem[\protect\citeauthoryear{Mohri, Sivek, and Suresh}{Mohri
  et~al\mbox{.}}{2019}]%
        {Mohri2019}
\bibfield{author}{\bibinfo{person}{Mehryar Mohri}, \bibinfo{person}{Gary
  Sivek}, {and} \bibinfo{person}{Ananda~Theertha Suresh}.}
  \bibinfo{year}{2019}\natexlab{}.
\newblock \showarticletitle{{Agnostic federated learning}}. In
  \bibinfo{booktitle}{{\em 36th International Conference on Machine Learning,
  ICML 2019}}, Vol.~\bibinfo{volume}{2019-June}.
  \bibinfo{publisher}{International Machine Learning Society (IMLS)},
  \bibinfo{pages}{8114--8124}.
\newblock
\showISBNx{9781510886988}
\showeprint[arxiv]{1902.00146}
\showURL{%
\url{http://arxiv.org/abs/1902.00146}}


\bibitem[\protect\citeauthoryear{Ruder}{Ruder}{2017}]%
        {Ruder2017}
\bibfield{author}{\bibinfo{person}{Sebastian Ruder}.}
  \bibinfo{year}{2017}\natexlab{}.
\newblock \showarticletitle{{An Overview of Multi-Task Learning in Deep Neural
  Networks}}.
\newblock  (\bibinfo{date}{jun} \bibinfo{year}{2017}).
\newblock
\showeprint[arxiv]{1706.05098}
\showURL{%
\url{http://arxiv.org/abs/1706.05098}}


\bibitem[\protect\citeauthoryear{Ruder, Bingel, Augenstein, and
  S{\o}gaard}{Ruder et~al\mbox{.}}{2019}]%
        {Ruder2019}
\bibfield{author}{\bibinfo{person}{Sebastian Ruder}, \bibinfo{person}{Joachim
  Bingel}, \bibinfo{person}{Isabelle Augenstein}, {and} \bibinfo{person}{Anders
  S{\o}gaard}.} \bibinfo{year}{2019}\natexlab{}.
\newblock \showarticletitle{{Latent Multi-Task Architecture Learning}}.
\newblock \bibinfo{journal}{{\em Proceedings of the AAAI Conference on
  Artificial Intelligence\/}} \bibinfo{volume}{33}, \bibinfo{number}{01}
  (\bibinfo{date}{jul} \bibinfo{year}{2019}), \bibinfo{pages}{4822--4829}.
\newblock
\showISSN{2159-5399}
\showDOI{%
\url{https://doi.org/10.1609/aaai.v33i01.33014822}}
\showeprint[arxiv]{1705.08142}


\bibitem[\protect\citeauthoryear{Sener and Koltun}{Sener and Koltun}{2018}]%
        {NIPS2018_7334}
\bibfield{author}{\bibinfo{person}{Ozan Sener} {and} \bibinfo{person}{Vladlen
  Koltun}.} \bibinfo{year}{2018}\natexlab{}.
\newblock \showarticletitle{{Multi-task learning as multi-objective
  optimization}}. In \bibinfo{booktitle}{{\em Advances in Neural Information
  Processing Systems}}, Vol.~\bibinfo{volume}{2018-Decem}.
  \bibinfo{pages}{527--538}.
\newblock
\showISSN{10495258}
\showeprint[arxiv]{1810.04650}
\showURL{%
\url{http://papers.nips.cc/paper/7334-multi-task-learning-as-multi-objective-optimization}}


\bibitem[\protect\citeauthoryear{Seo, Back, Lee, Park, Kim, and Lee}{Seo
  et~al\mbox{.}}{2020}]%
        {Seo2020}
\bibfield{author}{\bibinfo{person}{Hogeon Seo}, \bibinfo{person}{Seunghyeok
  Back}, \bibinfo{person}{Seongju Lee}, \bibinfo{person}{Deokhwan Park},
  \bibinfo{person}{Tae Kim}, {and} \bibinfo{person}{Kyoobin Lee}.}
  \bibinfo{year}{2020}\natexlab{}.
\newblock \showarticletitle{{Intra- and inter-epoch temporal context network
  (IITNet) using sub-epoch features for automatic sleep scoring on raw
  single-channel EEG}}.
\newblock \bibinfo{journal}{{\em Biomedical Signal Processing and Control\/}}
  \bibinfo{volume}{61} (\bibinfo{date}{aug} \bibinfo{year}{2020}),
  \bibinfo{pages}{102037}.
\newblock
\showISSN{17468108}
\showDOI{%
\url{https://doi.org/10.1016/j.bspc.2020.102037}}


\bibitem[\protect\citeauthoryear{Shokri and Shmatikov}{Shokri and
  Shmatikov}{2015}]%
        {10.1145/2810103.2813687}
\bibfield{author}{\bibinfo{person}{Reza Shokri} {and} \bibinfo{person}{Vitaly
  Shmatikov}.} \bibinfo{year}{2015}\natexlab{}.
\newblock \showarticletitle{{Privacy-preserving deep learning}}. In
  \bibinfo{booktitle}{{\em Proceedings of the ACM Conference on Computer and
  Communications Security}}, Vol.~\bibinfo{volume}{2015-Octob}.
  \bibinfo{publisher}{Association for Computing Machinery},
  \bibinfo{address}{New York, New York, USA}, \bibinfo{pages}{1310--1321}.
\newblock
\showISBNx{9781450338325}
\showISSN{15437221}
\showDOI{%
\url{https://doi.org/10.1145/2810103.2813687}}


\bibitem[\protect\citeauthoryear{Smith, Chiang, Sanjabi, and Talwalkar}{Smith
  et~al\mbox{.}}{2017}]%
        {NIPS2017_7029}
\bibfield{author}{\bibinfo{person}{Virginia Smith}, \bibinfo{person}{Chao~Kai
  Chiang}, \bibinfo{person}{Maziar Sanjabi}, {and} \bibinfo{person}{Ameet
  Talwalkar}.} \bibinfo{year}{2017}\natexlab{}.
\newblock \showarticletitle{{Federated multi-task learning}}. In
  \bibinfo{booktitle}{{\em Advances in Neural Information Processing Systems}},
  Vol.~\bibinfo{volume}{2017-Decem}. \bibinfo{pages}{4425--4435}.
\newblock
\showISSN{10495258}
\showURL{%
\url{http://papers.nips.cc/paper/7029-federated-multi-task-learning}}


\bibitem[\protect\citeauthoryear{S{\o}gaard and Goldberg}{S{\o}gaard and
  Goldberg}{2016}]%
        {Sagaard2016}
\bibfield{author}{\bibinfo{person}{Anders S{\o}gaard} {and}
  \bibinfo{person}{Yoav Goldberg}.} \bibinfo{year}{2016}\natexlab{}.
\newblock \showarticletitle{{Deep multi-task learning with low level tasks
  supervised at lower layers}}. In \bibinfo{booktitle}{{\em 54th Annual Meeting
  of the Association for Computational Linguistics, ACL 2016 - Short Papers}}.
  \bibinfo{pages}{231--235}.
\newblock
\showISBNx{9781510827592}
\showDOI{%
\url{https://doi.org/10.18653/v1/p16-2038}}


\bibitem[\protect\citeauthoryear{Vepakomma, Gupta, Dubey, and Raskar}{Vepakomma
  et~al\mbox{.}}{2019}]%
        {vepakommareducing}
\bibfield{author}{\bibinfo{person}{Praneeth Vepakomma},
  \bibinfo{person}{Otkrist Gupta}, \bibinfo{person}{Abhimanyu Dubey}, {and}
  \bibinfo{person}{Ramesh Raskar}.} \bibinfo{year}{2019}\natexlab{}.
\newblock \showarticletitle{{Reducing leakage in distributed deep learning for
  sensitive health data}}. In \bibinfo{booktitle}{{\em ICLR AI for social good
  workshop 2019}}. \bibinfo{pages}{1--6}.
\newblock


\bibitem[\protect\citeauthoryear{Vepakomma, Gupta, Swedish, and
  Raskar}{Vepakomma et~al\mbox{.}}{2018}]%
        {vepakomma2018split}
\bibfield{author}{\bibinfo{person}{Praneeth Vepakomma},
  \bibinfo{person}{Otkrist Gupta}, \bibinfo{person}{Tristan Swedish}, {and}
  \bibinfo{person}{Ramesh Raskar}.} \bibinfo{year}{2018}\natexlab{}.
\newblock \showarticletitle{{Split learning for health: Distributed deep
  learning without sharing raw patient data}}.
\newblock  (\bibinfo{date}{dec} \bibinfo{year}{2018}).
\newblock
\showeprint[arxiv]{1812.00564}
\showURL{%
\url{http://arxiv.org/abs/1812.00564}}


\bibitem[\protect\citeauthoryear{Wang, Yurochkin, Sun, Papailiopoulos, and
  Khazaeni}{Wang et~al\mbox{.}}{2020}]%
        {Wang2020}
\bibfield{author}{\bibinfo{person}{Hongyi Wang}, \bibinfo{person}{Mikhail
  Yurochkin}, \bibinfo{person}{Yuekai Sun}, \bibinfo{person}{Dimitris
  Papailiopoulos}, {and} \bibinfo{person}{Yasaman Khazaeni}.}
  \bibinfo{year}{2020}\natexlab{}.
\newblock \showarticletitle{{Federated Learning with Matched Averaging}}.
\newblock  (\bibinfo{date}{feb} \bibinfo{year}{2020}).
\newblock
\showeprint[arxiv]{2002.06440}
\showURL{%
\url{http://arxiv.org/abs/2002.06440}}


\bibitem[\protect\citeauthoryear{Yang, Liu, Chen, and Tong}{Yang
  et~al\mbox{.}}{2019b}]%
        {10.1145/3298981}
\bibfield{author}{\bibinfo{person}{Qiang Yang}, \bibinfo{person}{Yang Liu},
  \bibinfo{person}{Tianjian Chen}, {and} \bibinfo{person}{Yongxin Tong}.}
  \bibinfo{year}{2019}\natexlab{b}.
\newblock \showarticletitle{{Federated machine learning: Concept and
  applications}}.
\newblock \bibinfo{journal}{{\em ACM Transactions on Intelligent Systems and
  Technology\/}} \bibinfo{volume}{10}, \bibinfo{number}{2} (\bibinfo{date}{jan}
  \bibinfo{year}{2019}), \bibinfo{pages}{1--19}.
\newblock
\showISSN{21576912}
\showDOI{%
\url{https://doi.org/10.1145/3298981}}
\showeprint[arxiv]{1902.04885}


\bibitem[\protect\citeauthoryear{Yang and Hospedales}{Yang and
  Hospedales}{2017}]%
        {Yang2017}
\bibfield{author}{\bibinfo{person}{Yongxin Yang} {and}
  \bibinfo{person}{Timothy~M. Hospedales}.} \bibinfo{year}{2017}\natexlab{}.
\newblock \showarticletitle{{Deep multi-task representation learning: A tensor
  factorisation approach}}. In \bibinfo{booktitle}{{\em 5th International
  Conference on Learning Representations, ICLR 2017 - Conference Track
  Proceedings}}. \bibinfo{publisher}{International Conference on Learning
  Representations, ICLR}.
\newblock
\showeprint[arxiv]{1605.06391}
\showURL{%
\url{http://arxiv.org/abs/1605.06391}}


\bibitem[\protect\citeauthoryear{Yang, Chang, Zhang, and Liang}{Yang
  et~al\mbox{.}}{2019a}]%
        {Yang2019}
\bibfield{author}{\bibinfo{person}{Ziqi Yang}, \bibinfo{person}{Ee~Chien
  Chang}, \bibinfo{person}{Jiyi Zhang}, {and} \bibinfo{person}{Zhenkai Liang}.}
  \bibinfo{year}{2019}\natexlab{a}.
\newblock \showarticletitle{{Neural network inversion in adversarial setting
  via background knowledge alignment}}. In \bibinfo{booktitle}{{\em Proceedings
  of the ACM Conference on Computer and Communications Security}}.
  \bibinfo{publisher}{Association for Computing Machinery},
  \bibinfo{address}{New York, NY, USA}, \bibinfo{pages}{225--240}.
\newblock
\showISBNx{9781450367479}
\showISSN{15437221}
\showDOI{%
\url{https://doi.org/10.1145/3319535.3354261}}


\bibitem[\protect\citeauthoryear{Zhu and Jin}{Zhu and Jin}{2020a}]%
        {8744465}
\bibfield{author}{\bibinfo{person}{Hangyu Zhu} {and} \bibinfo{person}{Yaochu
  Jin}.} \bibinfo{year}{2020}\natexlab{a}.
\newblock \showarticletitle{{Multi-Objective Evolutionary Federated Learning}}.
\newblock \bibinfo{journal}{{\em IEEE Transactions on Neural Networks and
  Learning Systems\/}} \bibinfo{volume}{31}, \bibinfo{number}{4}
  (\bibinfo{date}{apr} \bibinfo{year}{2020}), \bibinfo{pages}{1310--1322}.
\newblock
\showISSN{21622388}
\showDOI{%
\url{https://doi.org/10.1109/TNNLS.2019.2919699}}
\showeprint[arxiv]{1812.07478}


\bibitem[\protect\citeauthoryear{Zhu and Jin}{Zhu and Jin}{2020b}]%
        {Zhu2020}
\bibfield{author}{\bibinfo{person}{Hangyu Zhu} {and} \bibinfo{person}{Yaochu
  Jin}.} \bibinfo{year}{2020}\natexlab{b}.
\newblock \showarticletitle{{Multi-Objective Evolutionary Federated Learning}}.
\newblock \bibinfo{journal}{{\em IEEE Transactions on Neural Networks and
  Learning Systems\/}} \bibinfo{volume}{31}, \bibinfo{number}{4}
  (\bibinfo{date}{apr} \bibinfo{year}{2020}), \bibinfo{pages}{1310--1322}.
\newblock
\showISSN{21622388}
\showDOI{%
\url{https://doi.org/10.1109/TNNLS.2019.2919699}}
\showeprint[arxiv]{1812.07478}


\bibitem[\protect\citeauthoryear{Zhu, Liu, and Han}{Zhu et~al\mbox{.}}{2019}]%
        {NIPS2019_9617}
\bibfield{author}{\bibinfo{person}{Ligeng Zhu}, \bibinfo{person}{Zhijian Liu},
  {and} \bibinfo{person}{Song Han}.} \bibinfo{year}{2019}\natexlab{}.
\newblock \showarticletitle{{Deep Leakage from Gradients}}.
\newblock In \bibinfo{booktitle}{{\em Advances in Neural Information Processing
  Systems 32}}, \bibfield{editor}{\bibinfo{person}{H~Wallach},
  \bibinfo{person}{H~Larochelle}, \bibinfo{person}{A~Beygelzimer},
  \bibinfo{person}{F~d$\backslash$textquotesingle Alch{\'{e}}-Buc},
  \bibinfo{person}{E~Fox}, {and} \bibinfo{person}{R~Garnett}} (Eds.).
  \bibinfo{publisher}{Curran Associates, Inc.}, \bibinfo{pages}{14774--14784}.
\newblock
\showURL{%
\url{http://papers.nips.cc/paper/9617-deep-leakage-from-gradients.pdf}}


\end{thebibliography}

\end{document}